\documentclass{article}
\usepackage{spconf,amsmath,graphicx,color,multirow, url}

\usepackage{paralist}

\title{Investigating Gated Recurrent Networks for Speech Synthesis}
\name{\parbox{6in}{\centering Zhizheng Wu \qquad Simon King}}
\address{The Centre for Speech Technology Research, University of Edinburgh, United Kingdom \\ {\small zhizheng.wu@ed.ac.uk}}
\begin{document}


%
\maketitle
\begin{abstract}
Recently, recurrent neural networks (RNNs) as powerful sequence models have re-emerged as a potential acoustic model for statistical parametric speech synthesis (SPSS). The long short-term memory (LSTM) architecture is particularly attractive because it addresses the vanishing gradient problem in standard RNNs, making them easier to train. Although recent studies have demonstrated that LSTMs can achieve significantly better performance on SPSS than deep feed-forward neural networks, little is known about why. Here we attempt to answer two questions: a) why do LSTMs work well as a sequence model for SPSS; b) which component (e.g., input gate, output gate, forget gate) is most important. We present a visual analysis alongside a series of experiments, resulting in a proposal for a simplified architecture. The simplified architecture has significantly fewer parameters than an LSTM, thus reducing generation complexity considerably without degrading quality.
\end{abstract}


\section{Introduction}
\label{sec:intro}

Statistical parametric speech synthesis (SPSS) has quite steadily advanced in naturalness in the past decade, as witnessed by the series of Blizzard Challenges~\cite{King2014_loquens}. However, the quality of synthetic speech produced by SPSS is still far below that of the natural human speech, and cannot compete with the best unit selection systems, which concatenate waveforms~\cite{hunt1996unit}. As suggested in~\cite{zen2009statistical}, acoustic modelling, which captures the complex relationship between linguistic and acoustic representations, is a key limiting factor and is the focus of this work.

\vspace{-3mm}
\subsection{Relation to prior work}
\vspace{-2mm}

Neural networks have re-emerged as a potential powerful acoustic model for SPSS. In~\cite{ze2013statistical,lu2013combining, qian2014training,hu2015fusion,cassia2015perceptual}, feed-forward neural networks are employed to map a linguistic representation derived from input text directly to acoustic features. In~\cite{kang2013multi}, a deep belief network (DBN) was used to model the relationship between linguistic and acoustic representations jointly. In~\cite{zen2014deep} and~\cite{uria2015modelling}, mixture density networks (MDNs) and real-valued neural autoregressive density estimators (RNADEs) were proposed, respectively, to predict acoustic feature distributions given input linguistic features. These various implementations can be viewed as a replacement of the decision tree in HMM-based speech synthesis; they map linguistic features to acoustic features frame by frame through multiple hidden layers. However, the temporal sequence nature of speech is not explicitly modelled in the network architectures.

To include temporal constraints, we proposed to include contextual information by stacking low-dimensional bottleneck features from multiple consecutive frames~\cite{wu2015deep}. Still in the DNN framework, minimum trajectory error training~\cite{wu2015minimum} or sequence error training criterion~\cite{fan2015sequence} have been proposed to minimise the utterance-level trajectory error rather than the frame-by-frame error. On the other hand, recurrent neural networks (RNNs) directly and elegantly include temporal information in the network architecture, making them attractive for modelling speech parameter trajectories. In~\cite{chen1998rnn}, a standard RNN was employed to predict prosodic information for speech synthesis. In~\cite{achanta2015investigation}, two variants on standard RNNs, the Elman RNN and clockwork RNN, were investigated for speech synthesis. 

The most widely used recurrent network in speech processing applications is the long short-term memory (LSTM) architecture. Because the LSTM addresses the vanishing gradient problem of the standard RNN, it is easier to train. In~\cite{fernandez2014prosody}, an LSTM was employed to model the F0 contour. In~\cite{fan2014rnn}, a bidirectional LSTM was employed to map a sequence of linguistic features to the corresponding sequence of acoustic features. In~\cite{zen2015unidirectional}, an LSTM with a recurrent output layer was proposed to perform sequence mapping from linguistic to acoustic representations. These studies all formulate SPSS as sequence-to-sequence mapping and all demonstrate the effectiveness of LSTMs. However, LSTM architecture seems rather ad-hoc and it is not obvious what its various components are actually contributing to performance.

This raises at least two questions that have not been answered in previous studies: a) \emph{how exactly does the LSTM architecture model a speech parameter sequence}; b) \emph{which components of the LSTM architecture are important, and which could be discarded}. Answers to these  questions may suggest better and perhaps simpler recurrent network architectures.

\vspace{-3mm}
\subsection{The novelty of this work}
\vspace{-2mm}

We attempt to reach a better understanding of the ``black-box'' LSTM architecture and our findings lead us to propose a simplified architecture for speech modelling.

First, we give an analysis of the forget gate and memory cell in the LSTM architecture. Specifically, we visualise the activation of the forget gate to understand when the forget gate resets the memory cell state, and how the forget gate relates to speech structure. We analyse how the cell state correlates with the trajectory to be predicted. These visualisations enable us to understand how LSTMs model the temporal structure in  speech synthesis. \emph{To the best of our knowledge, this is the first attempt to visually analyse the LSTM architecture in predicting a speech parameter sequence}.

Second, we analyse the importance of each LSTM component for speech synthesis and propose a simplified architecture. The analysis is done empirically with several variants of the LSTM. Each removes a different component of the vanilla LSTM. The analysis was inspired by the studies in~\cite{greff2015lstm,jozefowicz2015empirical}, and we focus on the speech synthesis application. Based on this analysis, we present a simplified architecture, which only has the forget gate. The simplified architecture has significantly fewer parameters than the vanilla LSTM, and so reduces the computational cost of generation considerably without degrading the quality of the synthesised speech.

\vspace{-2mm}
\section{Long Short-Term Memory}
\vspace{-3mm}

Standard RNNs are hard to train due to the well-known vanishing or exploding gradient problems~\cite{bengio1994learning,pascanu2012difficulty}. To address the vanishing gradient problem, the LSTM architecture was proposed, the basic idea of which was presented in~\cite{hochreiter1997long}. The most commonly used architecture was described in~\cite{graves2005framewise}, and is formulated as,
\begin{align}
\mathbf{i}_{t} &= \delta(\mathbf{W}^{\mathrm{i}}\mathbf{x}_{t} + \mathbf{R}^{\mathrm{i}} \mathbf{h}_{t-1} + \mathbf{p}^{\mathrm{i}} \odot \mathbf{c}_{t-1} + \mathbf{b}^{\mathrm{i}} ) \nonumber\\
\mathbf{f}_{t} &= \delta(\mathbf{W}^{\mathrm{f}}\mathbf{x}_{t} + \mathbf{R}^{\mathrm{f}} \mathbf{h}_{t-1} + \mathbf{p}^{\mathrm{f}} \odot \mathbf{c}_{t-1} + \mathbf{b}^{\mathrm{f}} ) \nonumber\\
\mathbf{c}_{t} &= \mathbf{f}_{t} \odot \mathbf{c}_{t-1} + \mathbf{i}_{t} \odot g(\mathbf{W}^{\mathrm{c}}\mathbf{x}_{t} + \mathbf{R}^{\mathrm{c}} \mathbf{h}_{t-1} + \mathbf{b}^{\mathrm{c}}) \nonumber\\
\mathbf{o}_{t} &= \delta(\mathbf{W}^{\mathrm{o}}\mathbf{x}_{t} + \mathbf{R}^{\mathrm{o}} \mathbf{h}_{t-1} + \mathbf{p}^{\mathrm{o}} \odot \mathbf{c}_{t} + \mathbf{b}^{\mathrm{o}} ) \nonumber\\
\mathbf{h}_{t} &= \mathbf{o}_{t} \odot g(\mathbf{c}_{t})\nonumber
\end{align}
In these formulations, $\mathbf{i}_{t}$, $\mathbf{f}_{t}$, $\mathbf{c}_{t}$, $\mathbf{o}_{t}$, and $\mathbf{h}_{t}$ are the input gate, forget gate, cell state, output gate and block output at time instance $t$, respectively; $\delta(\cdot)$ and $g(\cdot)$ are the sigmoid and tangent activation functions, respectively; $\mathbf{x}_{t}$ is the input at time $t$; $\mathbf{W}^{\mathrm{*}}$, and $\mathbf{R}^{\mathrm{*}}$ are the weight matrices applied on input and recurrent hidden units, respectively; $\mathbf{p}^{*}$ and $\mathbf{b}^{*}$ are the peep-hole connections and biases, respectively; and $\odot$ means element-wise product. We will call this the \emph{vanilla LSTM}.

The central idea of the LSTM is the so-called memory cell $\mathbf{c}$ which maintains its state over time, and the gating units which are used to regulate the information flow into and out of the memory cell~\cite{greff2015lstm}. More specifically, the input gate can allow the input signal to adjust the cell state or prevent that (e.g., setting the input gate to zero); the output gate can allow the cell state to affect other neurons or block that; and the forget gate enables the cell to remember or forget its previous state. However, as discussed in~\cite{greff2015lstm,jozefowicz2015empirical}, the architecture might not be optimal for all the tasks, and the relative importance of each component is not at all clear.

\vspace{-2mm}
\section{Gated Recurrent Neural Networks}
\vspace{-3mm}
In this section, we present several variants of the LSTM and propose a simplified version that only has the forget gate; it therefore has significantly fewer parameters and lower computational cost. As these variants all share with the LSTM the concept of a memory cell with gates, we will call them \emph{gated recurrent neural networks}.

\vspace{-2mm}
\subsection{Four variants on the LSTM}
\vspace{-1mm}
To assess the importance of each component, we start with four variants of the LSTM architecture. Each removes one component from the LSTM architecture, so we can understand how much each component contributes to performance. The differences with the vanilla LSTM are:
\begin{compactitem}
	\item No Peep-holes (NPH):  Set $\mathbf{p}^{\mathrm{i}}$, $\mathbf{p}^{\mathrm{f}}$, $\mathbf{p}^{\mathrm{o}}$ to zero
	\item No input gate (NIG):  $\mathbf{i}_{t} = 1$
	\item No forget gate (NFG): $\mathbf{f}_{t} = 1$
	\item No output gate (NOG): $\mathbf{o}_{t} = 1$
\end{compactitem}
In the NFG variant, the past cell state will still contribute to the current cell state but without any controlling  or scaling by the forget gate. Note that, when removing the input, forget or output gates, the number of parameters is reduced.

\vspace{-2mm}
\subsection{Gated Recurrent Unit (GRU)}
\vspace{-1mm}
As an alternative to the LSTM, the Gated Recurrent Unit (GRU) architecture was proposed in~\cite{cho2014learning}. In~\cite{chung2014empirical}, the GRU was found to achieve better performance than the LSTM on some tasks. The GRU is formulated as:
\begin{align}
\mathbf{r}_{t} &= \delta(\mathbf{W}^{\mathrm{r}}\mathbf{x}_{t} + \mathbf{R}^{\mathrm{r}} \mathbf{h}_{t-1} + \mathbf{b}^{\mathrm{r}} ) \nonumber\\
\mathbf{z}_{t} &= \delta(\mathbf{W}^{\mathrm{z}}\mathbf{x}_{t} + \mathbf{R}^{\mathrm{z}} \mathbf{h}_{t-1} + \mathbf{b}^{\mathrm{z}} ) \nonumber\\
\tilde{\mathbf{h}}_{t} &= g(\mathbf{W}^{\mathrm{h}} \mathbf{x}_{t} + \mathbf{r}_{t} \odot (\mathbf{R}^{\mathrm{h}} \mathbf{h}_{t-1}) + \mathbf{b}^{\mathrm{h}}) \nonumber\\
\mathbf{h}_{t} &= \mathbf{z}_{t} \odot \mathbf{h}_{t-1} + (1 - \mathbf{z}_{t}) \odot \tilde{\mathbf{h}}_{t}\nonumber
\end{align}
From these formulae, we can observed that the GRU architecture is similar to LSTM but without a separate memory cell. The GRU does not use peep-hole connections and output activation functions, and combines the input and forget gates into an update gate $\mathbf{z}_{t}$ to balance between previous activation $\mathbf{h}_{t-1}$ and the candidate activation $\tilde{\mathbf{h}}_{t}$. The reset gate $\mathbf{r}_{t}$ allows it to forget the previous state.

\vspace{-2mm}
\subsection{Simplified LSTM (S-LSTM)}

As we will see in the experiments reported in the next section, the input gate, output gate and peep-hole connections can be removed without degrading speech synthesis performance significantly. Hence, we can propose an even simpler variant, that removes output gates and peep-hole connections, and replaces the input gate by the forget gate in the form of $1 - \mathbf{f}_{t}$. In this way, only the forget gate is retained. This simplest variant can be written as:
\begin{align}
\mathbf{f}_{t} &= \delta(\mathbf{W}^{\mathrm{f}}\mathbf{x}_{t} + \mathbf{R}^{\mathrm{f}} \mathbf{h}_{t-1} + \mathbf{b}^{\mathrm{f}} ) \nonumber\\
\mathbf{c}_{t} &= \mathbf{f}_{t} \odot \mathbf{c}_{t-1} + (1 - \mathbf{f}_{t}) \odot g(\mathbf{W}^{\mathrm{c}}\mathbf{x}_{t} + \mathbf{R}^{\mathrm{c}} \mathbf{h}_{t-1} + \mathbf{b}^{\mathrm{c}}) \nonumber\\
\mathbf{h}_{t} &= g(\mathbf{c}_{t})\nonumber
\end{align}
The simplified architecture is similar to the GRU, except that it uses a memory cell state. The cell state is controlled by the forget gate only, which trades off between past cell state and current block input. When the activation of forget gate is small, the cell state will mainly depend on the block input, otherwise it will mainly copy the past cell state.

\begin{figure}[!t]
  \begin{center}
    \includegraphics[width=70mm]{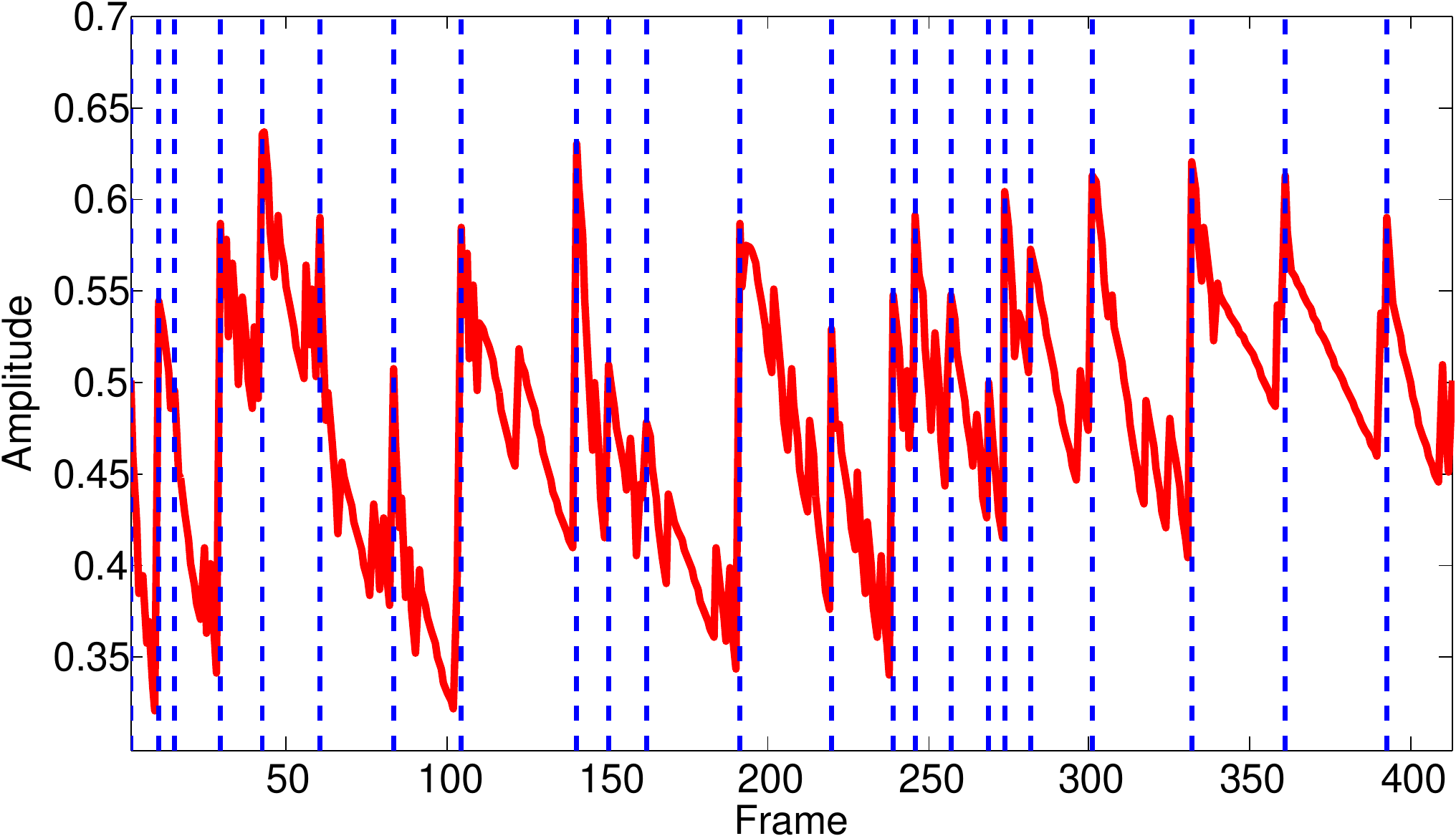}
	\vspace{-4mm}
    \caption{Averaged activations of all 256 forget gates as a function of time (in frames) are shown by the solid red line, with phoneme boundaries marked as dashed vertical blue lines.}
    \label{fig:forget_gate}
  \end{center}
\vspace{-6mm}
\end{figure}

\section{Experiments}
\vspace{-3mm}
\subsection{Experimental setup}
\vspace{-1mm}
A corpus from a British male speaker was employed in our experiments, divided into three subsets: training, development and testing (2400, 70 and 72 utterances). The sampling rate was 48 kHz, and we used the STRAIGHT vocoder~\cite{kawahara1999restructuring} to extract 60-dimensional Mel-Cepstral Coefficients (MCCs), 25 band aperiodicities (BAPs), and fundamental frequency ($F_{0}$) on log-scale, all at 5-ms frame step. All systems used the same acoustic features. $F_{0}$ was linearly interpolated before modelling and a binary voiced/unvoiced feature was used to record voicing information. Dynamic features for MCCs, BAPs and $F_0$ were also computed. The acoustic features were mean-variance normalised before modelling, and the mean and variance was restored at the generation time. At generation time, maximum likelihood parameter generation algorithm~\cite{tokuda2000speech} was applied to smooth parameter trajectories.

All systems used the same input linguistic features comprising 601 features. 592 of these are binary features derived from linguistic context, such as quin-phone identities, part-of-speech, positional information of phoneme, syllable, word and phrase, and the number of syllables, words and phrases, etc. The remaining 9 numerical features capture frame position information, e.g., frame position in HMM state and phoneme. Linguistic features were normalised to [0.01 0.99] before modelling.

In all RNNs, we employed a three-layer feed-forward neural network at the bottom. On top of the feed-forward layers, we used the gated recurrent neural networks.
The bottom feed-forward layers were intended to act as feature extraction layers, with  512 hidden units using tangent activation function in each layer. All RNN implementations used 256 units (e.g., LSTM blocks) in the recurrent layer. Hyperparameters for each system were optimised on the development set. We fixed the momentum, and only tuned learning rates.

\begin{figure}[!t]
  \begin{center}
    \includegraphics[width=70mm]{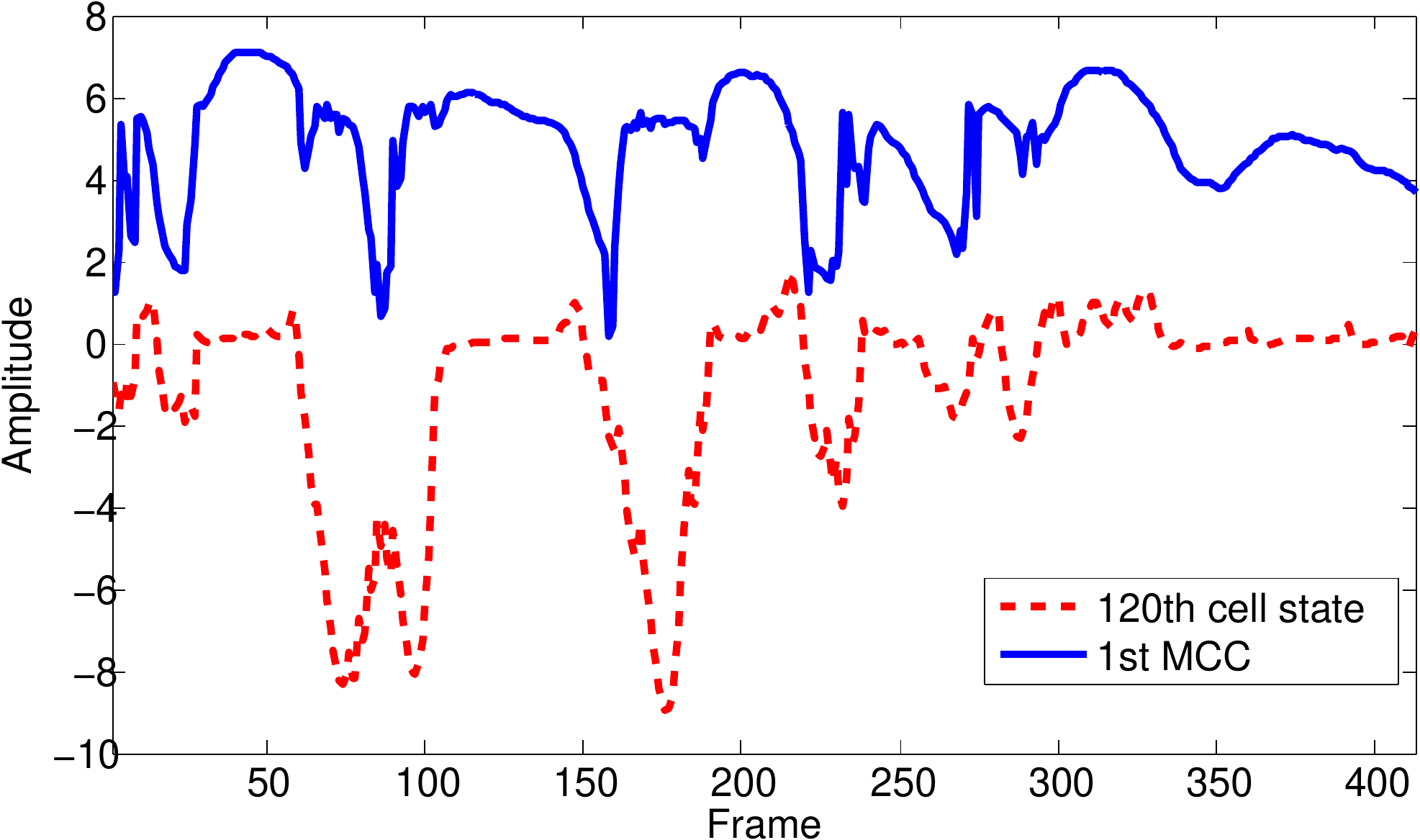}
	\vspace{-4mm}
    \caption{Comparison between the first MCC trajectory and the $120^{\mathrm{th}}$ cell state (vertically offset for clarity).}
    \label{fig:cell_state}
  \end{center}
\vspace{-6mm}
\end{figure}

\begin{table*}[!t]
\scriptsize
\center
\caption{Subjective preference scores (\%). $p < 0.01$ indicates significant difference between the two systems.}
\label{tbl:subjective_results}
\begin{tabular}{|l|cccccccc|c|}
  \hline
           &     LSTM     &  NIG          &  NOG        &  NFG          &    NPH             &   GRU           & S-LSTM          &  Neutral       &  p-value      \\
  \hline \hline
	1      &     46.7     &  42.3         &  -          &  -            &   -                &   -             &  -              &  11.0          &  0.1582       \\
	2      &     40.7     &  -            &  47.8       &  -            &   -                &   -             &  -              &  11.5          &  0.0649       \\
	3      & \textbf{74.3}&  -            &  -          & \textbf{20.3} &   -                &   -             &  -              &  5.4           &  $<10^{-8}$   \\
	4      &     44.0     &  -            &  -          &  -            &   45.3             &   -             &  -              &  10.7          &  0.7820       \\
	5      &     47.8     &  -            &  -          &  -            &   -                &   42.7          &  -              &  9.5           &  0.1086       \\
	6      &     46.5     &  -            &  -          &  -            &   -                &   -             &  45.3           &  8.2           &  0.7948      \\

	7      &     -        &  46.0         &  -          &  -            &   -                &   -             &  46.8           &  7.2           &  0.8305      \\
	8      &     -        &  -            &  48.0       &  -            &   -                &   -             &  44.3           &  7.7           &  0.3828      \\
	9      &     -        &  -            &  -          &  -            &   51.7             &   -             &  40.5           &  7.8           &  0.0018      \\
	10     &     -        &  -            &  -          &  -            &   -                &   48.0          &  45.0           &  7.0           &  0.4703      \\
  \hline
\end{tabular}
\end{table*}

\vspace{-3mm}
\subsection{Analysis of LSTM}
\vspace{-2mm}
We first visualised the forget gate and cell state, which are thought to be the two most important components in modelling long-term temporal structure. The averaged activations (over the 256 units) of the forget gate as a function of the frame index is presented in Fig.~\ref{fig:forget_gate}. The red solid line is the forget gates averaged activations; blue dashed lines show phoneme boundaries. It is clear that the peaks of the forget gate activation trajectory have a strong correspondence with the phoneme boundaries; within a phoneme, the contribution of past cell state decays linearly. The forget gate is capturing some important temporal structure of speech; this is not surprising, since the phoneme boundaries are explicitly represented in the input linguistic features.

\begin{table}[!h]
\scriptsize
\center
\caption{Objective measures. MCD: Mel-Cepstral Distortion. BAP: distortion of band aperiodicities. F0 RMSE is calculated on a linear scale. V/UV: voiced/unvoiced error. Note that the number of parameters listed is for the recurrent layer only and does not include the bottom three  feed-forward layers, which are the same size across all systems. The generation time is to generate all the 142 utterances in both development and testing sets.}
\label{tbl:objective_results}
\begin{tabular}{|l|cccc|cc|}
  \hline
                   &  MCD      &  BAP           &  F0 RMSE    &  V/UV         &   \# parameters    & generation   \\
                   &  (dB)     &  (dB)          &  (Hz)       &  (\%)         &                    & time (s)         \\
  \hline \hline
LSTM               &  \textbf{4.14}     &  1.95          &  8.96       &  4.15         &    788, 224        & 214               \\
NIG                &  4.18              &  1.95          &  9.10       &  4.15         &    591, 104        & 180               \\
NOG                &  \textbf{4.14}     &  \textbf{1.94} &\textbf{8.84}&  4.29         &    591, 104        & 167               \\
NFG                &  4.68              &  1.99          &  9.69       &  4.41         &    591, 104        & 174               \\
NPH                &  \textbf{4.14}     &  1.95          &  9.02       &  4.20         &    787, 456        & 180               \\
GRU                &  4.17              &  1.95          &  9.00       &  4.22         &    590, 592        & 159               \\
S-LSTM             &  4.19              &  1.95          &  8.87       & \textbf{4.14} &  \textbf{393, 728} & \textbf{154}               \\

  \hline
\end{tabular}
\vspace{-4mm}
\end{table}

The memory cell should maintains its state over time~\cite{greff2015lstm} and so could store the trend of the trajectory to be predicted.
To analyse the relationship between the cell states and the MCC trajectories, we computed the correlation between the cell states and the first MCC trajectory, and found that the $120^{\mathrm{th}}$ cell state has the highest correlation with the first MCC trajectory. The correlation is as high as 0.9. A comparison between them is presented in Fig.~\ref{fig:cell_state}, which shows that the cell state tracks the shape of the MCC trajectory.

\subsection{Objective results}
\vspace{-1mm}
Even though objective measures might not always correlate with human perception, they offer a way to tune the systems and roughly predict model performance. The objective results are in Table~\ref{tbl:objective_results}. Compared to LSTM, NIG, NOG and NPH all achieve similar objective distortion, with considerably fewer parameters and lower generation time: the input gate, output gate and peep-hole connections are not necessary. The NFG system increases distortion considerably: the forget gate is important. This finding is consistent with~\cite{greff2015lstm}.

The GRU system achieves similar performance to the LSTM system: even though it has even fewer parameters, it performs as well as NIG, NOG or NPH. This is also consistent with studies on other tasks~\cite{chung2014empirical,jozefowicz2015empirical}. Although S-LSTM slightly increases MCD distortion from 4.14 dB to 4.19 dB compared to LSTM, it achieves similar performance on the other measures. The S-LSTM has about half the number of parameters in its recurrent layer compared to the LSTM, and reduces generation time from 214 seconds to 154 seconds. The generation time is the total time to generate all the 142 utterances in both development and testing sets.

In summary, the S-LSTM has the smallest number of parameters and achieves the fastest generation, whilst achieving similar objective results to the LSTM and GRU architectures.

\subsection{Subjective results}
\vspace{-1mm}
Subjective preference tests were conducted using 30 paid native English speakers. Each listener was asked to listen 20 pairs of synthesised utterances. The sentence was the same in both items within a pair, and was randomly selected from the 72 test sentences\footnote{Samples are available at: \url{http://homepages.inf.ed.ac.uk/zwu2/demo/icassp16/lstm.html}}. For each pair, the listener was asked to decide which one sounded more natural; a ``neutral'' option was allowed if the listener had no preference.

Preference results are in Table~\ref{tbl:subjective_results}. Comparing against the LSTM system, all the systems except NFG show no significant difference in  preference. 

The NFG system achieves only a 20.3\% preference score when paired against the LSTM which is preferred 74.3\% of the time. As with the objective results in Table~\ref{tbl:objective_results}, we conclude that the forget gate is the only critical component in the LSTM architecture; the input gate, output gate and peep-hole connections can be omitted.

We also compares the proposed S-LSTM system against with all other systems (except NFG, since it is worse than LSTM). Consistent with the objective results, the subjective results also demonstrate that S-LSTM is as good as any other systems.
\vspace{-4mm}

\section{Conclusions}
\vspace{-3mm}
We have analysed the forget gate and cell state of the LSTM architecture, and examined the performance of several variants of LSTM. We conclude that:
\begin{compactitem}
	\item The forget gate can learn the temporal structure of speech; its activations have a high correspondence with phone boundaries.
	\item The memory cell maintains a state over time, which matched the shape of the trajectory to be predicted.
	\item For this task, the forget gate is the only critical component of the LSTM;  other components can be omitted with no reduction in naturalness. 
\end{compactitem}

From these results, we propose a simplified LSTM architecture that only uses the critical forget gate. The simplified LSTM has significantly fewer parameters than the vanilla LSTM, but achieves similar performance in both objective and subjective evaluations.

\vspace{1mm}
\ninept
{\footnotesize \textbf{Acknowledgements:} This research was supported by EPSRC Programme Grant EP/I031022/1, Natural Speech Technology (NST). The NST research data collection may be accessed at \newline http://datashare.is.ed.ac.uk/handle/10283/786.}

\bibliographystyle{IEEEbib}
\ninept
\bibliography{refs}

\end{document}